\documentclass[runningheads]{llncs}
\usepackage[T1]{fontenc}
\usepackage{subcaption}
\usepackage{graphicx}
\usepackage{amsmath,amssymb,amsfonts}
\usepackage{makecell}
\usepackage{multirow}
\usepackage{booktabs}
\usepackage{comment}
\usepackage{hyperref}
\usepackage{marvosym}
\begin{document}
%

\title{Region-specific Risk Quantification for Interpretable Prognosis of COVID-19}

\titlerunning{Regional Interpretable Prognosis}
%
\author{Zhusi Zhong\inst{1,2,3} \and 
Jie Li\inst{1} \and 
Zhuoqi Ma\inst{1,2,3} \and 
Scott Collins\inst{2,3} \and
Harrison Bai\inst{4} \and \\
Paul Zhang\inst{5} \and
Terrance Healey\inst{2,3} \and
Xinbo Gao\inst{1} \and \\
Michael K. Atalay\inst{2,3} \and
Zhicheng Jiao \inst{2,3}\textsuperscript{(\Letter)}}
\authorrunning{Z. Zhong et al.}
%
\institute{School of Electronic Engineering, Xidian University, Xi’an, China \and
Warren Alpert Medical School, Brown University, Providence, USA \and
Department of Diagnostic Imaging, Rhode Island Hospital, Providence, USA \email{zhicheng\_jiao@brown.edu}\\ \and
Department of Radiology and Radiological Sciences, Johns Hopkins University School of Medicine, Baltimore, USA \and
Hospital of the University of Pennsylvania, Pennsylvania, USA
} 

\maketitle            

\begin{abstract}
The COVID-19 pandemic has strained global public health, necessitating accurate diagnosis and intervention to control disease spread and reduce mortality rates. This paper introduces an interpretable deep survival prediction model designed specifically for improved understanding and trust in COVID-19 prognosis using chest X-ray (CXR) images. By integrating a large-scale pretrained image encoder, Risk-specific Grad-CAM, and anatomical region detection techniques, our approach produces regional interpretable outcomes that effectively capture essential disease features while focusing on rare but critical abnormal regions. Our model's predictive results provide enhanced clarity and transparency through risk area localization, enabling clinicians to make informed decisions regarding COVID-19 diagnosis with better understanding of prognostic insights. We evaluate the proposed method on a multi-center survival dataset and demonstrate its effectiveness via quantitative and qualitative assessments, achieving superior C-indexes (0.764 and 0.727) and time-dependent AUCs (0.799 and 0.691). These results suggest that our explainable deep survival prediction model surpasses traditional survival analysis methods in risk prediction, improving interpretability for clinical decision making and enhancing AI system trustworthiness.

\end{abstract}

\section{Introduction}
The COVID-19 pandemic has strained global public health with rapid viral transmission leading to widespread infection and depleted healthcare resources. Immediate diagnosis and intervention are crucial for controlling disease spread and reducing mortality rates. Medical imaging, including computerized tomography (CT) and chest X-ray (CXR) scans, aids in expedited COVID-19 screening, often revealing pulmonary manifestations like ground-glass opacities. In current clinical practice, radiologists are required to manually inspect delineated anatomical regions in CXRs both normal and abnormal findings \cite{goergen2013evidence}. Radiologists face challenges in meeting the demands of their daily workload. Moreover, there is a shortage of specialized radiologists which has been exacerbated by the pandemic \cite{rimmer2017radiologist}. AI research in radiology has seen significant progress in diagnostics and decision making, with deep survival prediction models assessing the impact of symptom onset to diagnosis time on patient outcomes and identifying those requiring earlier consultation to prevent overwhelming health services \cite{gao2021risk}. \par 

However, the current challenge faced by deep survival prediction models lies in their inherent lack of interpretability, often referred to as the "black box" problem \cite{geis2019ethics,guidotti2018survey}. While these survival analysis models such as Random survival forests (RSF) \cite{rsf}, Cox Proportional Hazards (CoxPH) \cite{coxmodel}, even with certain deep learning algorithms \cite{deepsurv}, frequently exhibit high predictive accuracy, they lack transparency in elucidating the reasoning behind their predictions. In survival prediction, opacity hinders its practical implementation due to the following reasons: (1) reluctance from healthcare professionals to trust non-transparent models; (2) ethical concerns over fairness and bias prevention; (3) difficulties in clinical decision making without understanding predictions; (4) regulatory compliance challenges as clear explanations for predictions may be lacking, which compromises adherence to transparency and accountability requirements. Existing techniques are mainly based on class activation mapping (CAM) \cite{zhou2016learning} that have the capability to localize the coarse regions either by extracting information using global average pooling or any layer of choice using gradient information, such as Gradient-weighted Class Activation Mapping (Grad-CAM) \cite{selvaraju2017grad}. \par 
 
To ensure the model focuses adequately and accurately captures these rare yet crucial abnormal regions, this study proposes a survival prediction model designed to produce regionally interpretable outcomes. Leveraging a large-scale pretrained backbone, the proposed model effectively captures COVID-19 risk features from CXR images. Moreover, the predictive outcomes from our model offer enhanced clarity and transparency through an emphasis on both disease region localization and risk categorization. Subsequently, we conducted experiments on a multicenter dataset to quantitatively and qualitatively verify the effectiveness of our model. Our contributions include: (1) Proposing an efficient survival analysis model based on large-scale pretraining, demonstrating its efficacy in COVID-19 survival analysis tasks. (2) Utilizing Grad-CAM in our survival prediction model to generate survival risk attention maps, facilitating global risk localization. (3) Additionally, offering regionally interpretable predictions, enabling the identification of anatomical regions' risk levels and providing clinicians with comprehensible prognostic insights.

\begin{figure}
\centering
\includegraphics[scale=.14]{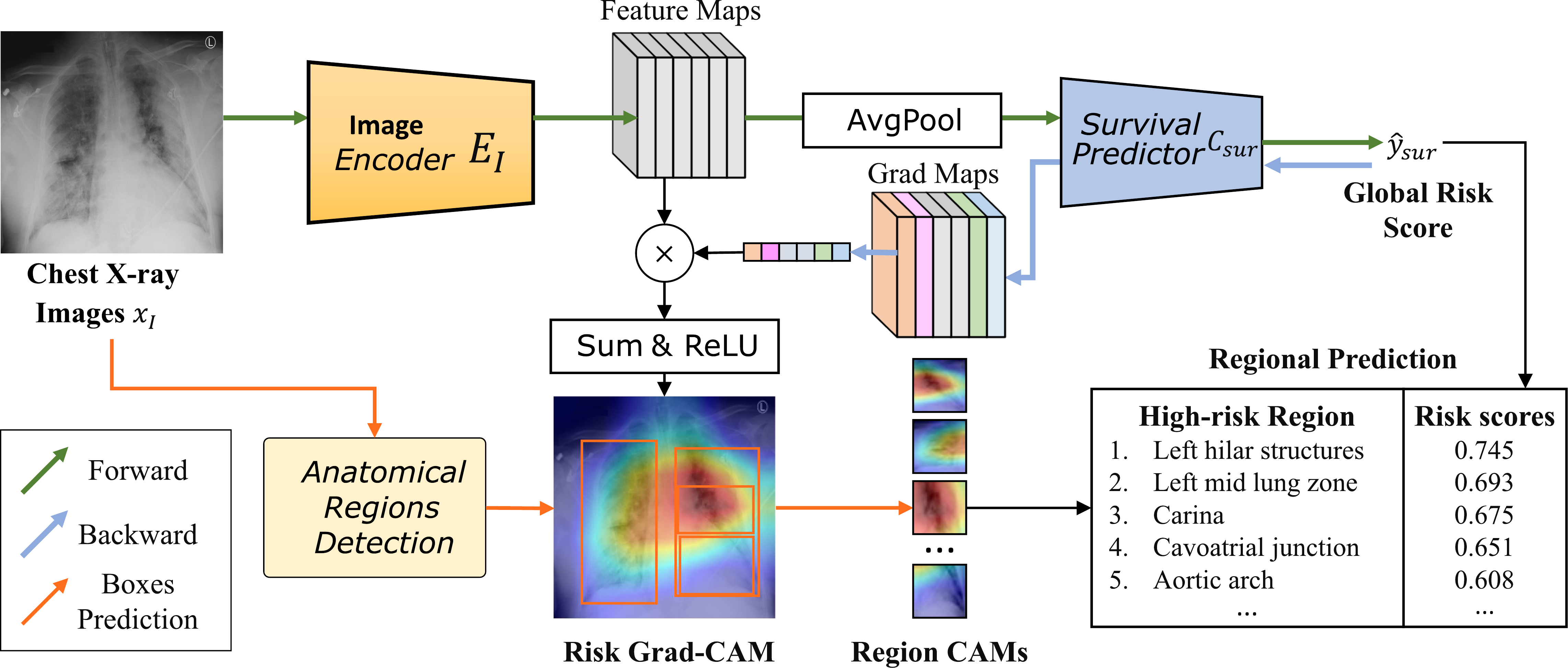}
\caption{Overview of the Survival Prediction (SP) model with regional interpretable prediction. The SP model first generates risk scores based on input CXR (Green). The risk Grad-CAM is calculated by backpropagating the global risk scores to locate disease areas (Blue). A branch detects the 29 anatomical bounding boxes for computing regional risk Grad-CAM and scores, arranging the region names in risk levels for a interpretable and understandable output to healthcare professionals (Orange).
} \label{fig1}
\end{figure}

\section{Methods}
\subsection{COVID-19 Survival Prediction Model}
We conducted COVID-19 survival prediction on CXR images with an efficient pretrained encoder, represented by green flow shown in Fig. \ref{fig1}. Specifically, we employ PRIOR—a Medical Vision Language Pretraining model \cite{cheng2023prior} that has undergone large-scale pretraining with conditional reconstruction tasks for both vision and language representation on the MIMIC-CXR dataset \cite{johnson2019mimic}. To extract image features, we utilize ResNet50 structure in PRIOR as our image encoder $E_I$, which learned visual features through pretraining on a large-scale CXR dataset.

Given an image $x_I$, the visual features are extracted from last Res-Layer and denoted as $f_I = E_I(x_I)$, $\in R^{{C} \times {H} \times {W}}$. The survival predictor $C_{sur}$ is designed to discern the patients' survival level from the average pooled visual features. The computation process of survival outcomes $\hat{y}_{sur}$ is as follows:
\begin{equation}
\hat{y}_{sur} = C_{sur} \left ( \mathrm{AvgPool} \left (E_I(x_I) \right )\right )
\end{equation}
The survival prediction model is trained to distinguish between patients with critical disease and those without, utilizing the CoxPH loss function \cite{deepsurv}, which is adept at handling right-censored survival data and is formulated as:
\begin{equation}
 L_{\mathrm{CoxPH}}= -\frac{1}{N_{Y_{e}=1}} \sum_{i: y_{e}^{i}=1} \left(\hat{y}^{i}_{sur} - \log \sum_{j:y_{t}^{j}>y_{t}^{i}}e^{\hat{y}^{j}_{sur}}\right) 
\end{equation}
where $\hat{y}^{i}_{sur}$ represents the predicted risk of disease progression for the $i$-th patient. Survival labels $y^{i}_{t}$ and $y^{i}_{e}$ are donated as the survival time and the censorship flag, respectively. $y^{i}_{t}$ is the days of diagnosis to death (for $y^{i}_{e}=1$) or patient censored (for $y^{i}_{e}=0$), $N_{Y_{e}=1}$ refers to the number of patients with an observable event.

\subsection{Risk-specific Grad-CAM}
Grad-CAM \cite{selvaraju2017grad} is a technique commonly utilized for elucidating significant image features employed by neural network across various tasks, including image classification. In the context of survival prediction, understanding the image regions attended by the model during prediction can yield valuable insights into underlying disease processes and prognostic factors within visual space. We compute the attention enhanced risk distribution map of the predicted outcome $\hat{y}_{sur}$ with respect to the risk-specific activation of visual feature maps.\par
Given $f_I$ extracted from the input image $x_I$ and the corresponding survival prediction score $\hat{y}_{sur}$, the gradients of $\hat{y}_{sur}$ to $f_I$ of the last convolutional layer are calculated. These gradients, flowing through backpropagation, are then globally average-pooled to derive the neuron importance weights $\alpha_I^c$:
\begin{equation}
\alpha_I^c = \frac{1}{Z} \sum_i \sum_j \frac{\partial \hat{y}_{sur}}{\partial f_I^c(i,j)}
\end{equation}
where $Z$ represents the normalization factor, and $i$ and $j$ denote spatial positions within the feature map. The $C$-dimensional weight vector $\alpha_I$ linearly captures the channel-wise 'importance' of the feature map to the global risk prediction.\par
The risk-specific Grad-CAM activation map ($L_{Risk-CAM}$) is computed by performing a channel-weighted sum of the feature maps $f_I$ using the computed gradients $\alpha_I$, followed by a rectified linear unit (ReLU) operation:
\begin{equation}
L_{Risk-CAM} = \mathrm{ReLU} \left( \sum_{c \in C} \alpha_I^c f_I^c \right)
\end{equation}
Specifically, the linear combination is chosen to focus solely on features positively influencing the risk predictions. These pixels, typically associated with higher disease relevance, often correspond to lesion areas contributing to increased risk. ReLU activation filters out negative pixels likely belonging to irrelevant categories in X-ray images, such as background and artificial artifacts. They are subsequently upsampled to match the spatial dimensions of the input image, facilitating direct visualization of a global and coarse level location contributing to the survival prediction.

\begin{figure}[!t]
\centering
\includegraphics[scale=0.14]{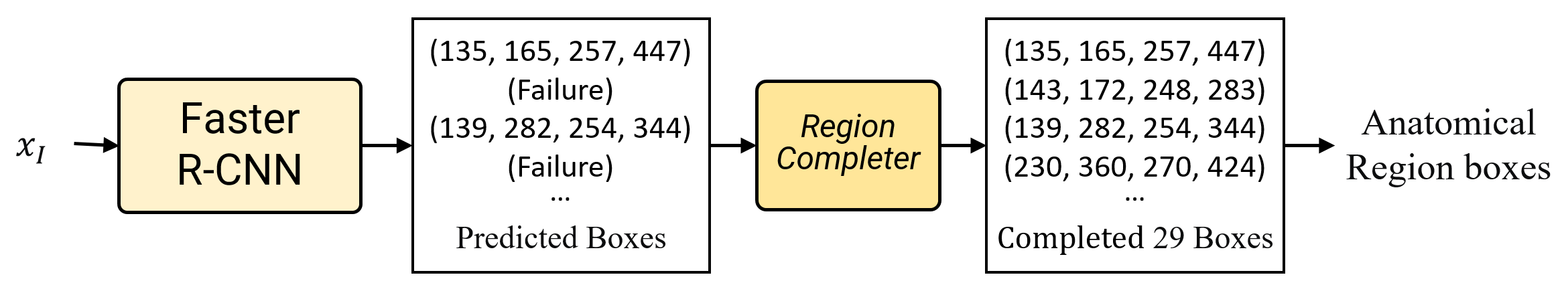}
\caption{The illustration of the anatomical region detector with Faster R-CNN and the proposed Region Completer, which corrects bounding box of the undetected regions with the learned spatial coordinate pattern.} \label{completer}
\end{figure}
\subsection{Anatomical region detection and completion} 
To improve global results' spatial readability, we align them to regional survival predictions with anatomical regions. The anatomical region detector is built on a Faster R-CNN model \cite{ren2015faster}, which was pre-trained on the Chest ImaGenome dataset \cite{wu2021chest}, to predicts 29 unique anatomical regions per case. Following standard procedure, the detector identifies well-detected and undetected regions based on probability scores. The well-detected region has highest probability score among the 29 classes, as well as the top-1 scores for all proposals. Otherwise, the region class does not achieve the highest score for at least one proposal, it is considered as undetected. \par

As illustrated in Fig. \ref{completer}, the Region Completer is a specialized and compact network, devised to rectify bounding boxes for undetected regions. The missing bounding boxes are estimated by leveraging the initial predictions generated by the Faster R-CNN. Completer functions as a regression network comprising three-layer MLPs, and is trained in the way of "Masked Autoencoders" \cite{he2022masked}. Specifically, during training, it predicts coordinates of several missing regions which are artificially randomly masked, thereby learning the spatial pattern of the 29 region coordinates. In the inference phase of region detection, the previously predicted coordinates utilized as fixed coordinates, potentially containing undetected regions. The relative distribution correlations between known and undetected regions are then utilized to infer the missing bounding boxes, thereby completing the bounding boxes for all 29 regions.

\subsection{Regional Risk Prediction}
Utilizing the completed bounding boxes, we delineate risk Grad-CAM and evaluate survival scores within each anatomical region. Within the Risk Grad-CAM, regional CAMs are extracted within each bounding box, as illustrated in Figure \ref{fig1}. Additionally, the computation of regional risk scores is predicated on the quantity of highlighted activation pixels within each anatomical region. The sum of pixel values within each region is normalized to determine the regional relative activation intensity. Multiplying these values by the global risk score yields the regional risk score. Furthermore, sorting the regional risk scores provides clinicians with a concise reference for prioritizing high-risk regions, thereby streamlining the manually double-check process and avoiding laborious verification of highly activated areas on the global Grad-CAMs. This enhances the interpretability of the predictive results, offering localized insights that allow clinicians to better understand the model's outcomes.

\section{Experiment}
\subsection{Datasets}
We validate the proposed survival prediction model and regional experiments on the multi-center survival dataset. A retrospective review was conducted to identify COVID-19 patients with imaging between March 2020 and July 2020 at Hospital (A) and the Hospital (B). The dataset A contains 1,021 front-view CXR and corresponding the survival labels, which include mortality event and the length of time from COVID-19 test to date of follow-up or mortality. We conducted our model evaluations in a multi-center setting to ensure robustness and generalizability. The dataset A was divided into training, validation, and testing sets, with a ratio of 7:1:2, maintaining ratio of survival events. The dataset B with 2,879 samples was used external testing. The study and collection of dataset have been approved by xxxxx protocol.

The Chest ImaGenome dataset \cite{wu2021chest} are used to train the anatomical region detector and completer. The dataset contains the automatically constructed scene graphs for the MIMIC-CXR \cite{johnson2019mimic} dataset, consisting of CXR images with corresponding free-text radiology reports. Each scene graph describes one frontal CXR and contains bounding box coordinates for 29 unique anatomical regions, as well as sentences describing each region if they exist in the corresponding radiology report. We use the split provided by the dataset: 166,512 training, 23,952 validation, and 47,389 test images.  

\subsection{Implementation Details}
All images were resized to 224$\times$224 while preserving the original aspect ratio, and normalized to zero mean and unit standard deviation. Color jitter, Gaussian noise, and affine transformations were applied as image data augmentations during training. The parameter of the image encoder with large-scale pretrain were fixed for survival prediction. A fully connected layer with Sigmoid acts as a risk classifier $C_{sur}$ for survival prediction. To train SP model, we set batch size as 1 and optimize with the $AdamW$ optimizer (a momentum of 0.9, a weight decay of 0.0005), and a learning rate of $10^{-5}$. While the maximum 200 epochs, we saved the parameters of best validation for inference. For region detection, we reload the checkpoint of RGRG \cite{tanida2023interactive_rgrg} to Faster R-CNN and trained the proposed region completer individually with a special setting (batch size: 2000, learning rate: $10^{-3}$). The model was implemented on PyTorch and pycox library \cite{kvamme2019time} for survival evaluation. Experiments were conducted on an Nvidia 3090 GPU.

\begin{table}[!t]
\scriptsize
\renewcommand{\arraystretch}{0.75}
    \centering
    \caption{\label{tab1} Performance comparison and ablation analysis of the proposed SP model. The larger values are with better performance. }
    \setlength{\tabcolsep}{1.8mm}{
    \begin{tabular}{c|c|c|c|c|c|c}
    \toprule[ 0.8 pt]
        & & & \multicolumn{2}{c|}{C-Index} & \multicolumn{2}{c}{T-AUC} \\ 
        Method & Pretrain & Fix $E_I$ & Testset A & Testset B & Testset A & Testset B\\ 
        \midrule [ 0.4 pt]
        RSF \cite{rsf} & - & - & 0.752 & 0.663 & 0.732 & 0.685 \\
        CoxPH \cite{coxmodel} & - & - & 0.687 & 0.680 & 0.701 & 0.612\\
        Ours & No & $\times$ & 0.614  & 0.547 & 0.642 & 0.529\\
        Ours & ImageNet & $\times$ & 0.736  & 0.638 & 0.739 & 0.610\\
        Ours & ImageNet & $\checkmark$ & 0.706  & 0.601 & 0.716 & 0.560\\
        Ours & PRIOR \cite{cheng2023prior} & $\times$ & 0.726  & 0.668 & 0.772 & 0.634\\
        Ours & PRIOR \cite{cheng2023prior} & $\checkmark$ & $\mathbf{0.764}$  & $\mathbf{0.727}$ & $\mathbf{0.799}$ & $\mathbf{0.691}$\\
    \bottomrule [ 0.8 pt] 
    \end{tabular}}
\end{table}

\begin{figure}[!t]
\centering
\includegraphics[scale=.20]{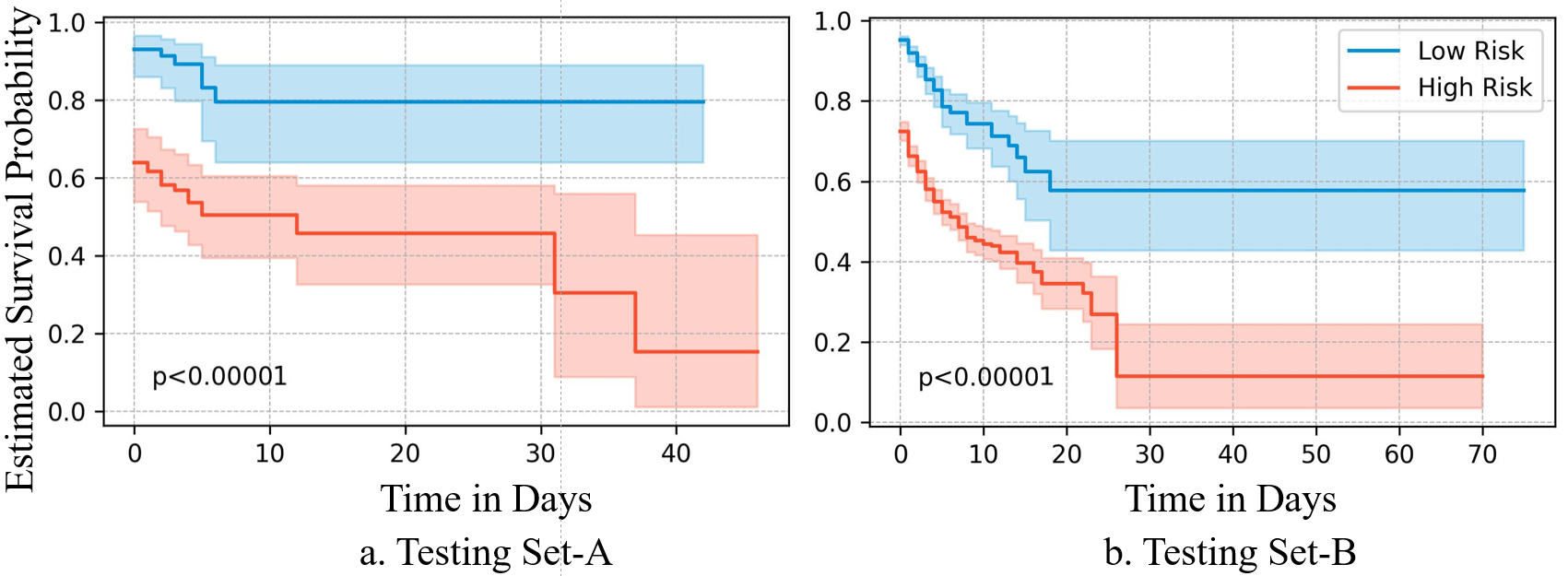}
\caption{Survival curves visualization of survival probabilities for patients of multi-center testing sets, classified into high and low risk groups. } \label{km}
\end{figure}

\subsection{Results}
Table \ref{tab1} highlights quantitative evaluations of traditional methods and our model using the Concordance Index (C-index) \cite{cindex} and time-dependent AUC (T-AUC) \cite{hung2010estimation}. Our proposed model demonstrates C-indexes of 0.764 and 0.727, surpassing survival analysis in both testing sets. With two days serving as critical time points of COVID-19 patient \cite{gao2021risk}, our model achieves T-AUC performance values of 0.799 and 0.691, reflecting superior capability at differentiating patients with/without death. The Kaplan-Meier (K-M) \cite{km} survival curve in Fig. \ref{km} confirms the effectiveness by demonstrating significant differences between high-risk and low-risk patient groups on predicted risk median, supported by $p$-values less than 0.0001 from a hypothesis test. Ablation studies on pretrained image encoder models reveal that large-scale CXR pretraining significantly enhances model generalization ability, particularly when using limited local data sets. Fixing the large-scale CXR encoder also results in superior performance, as overfitting to small local datasets leads to a performance decline.\par

\begin{figure}[!t]
\centering
\includegraphics[scale=.19]{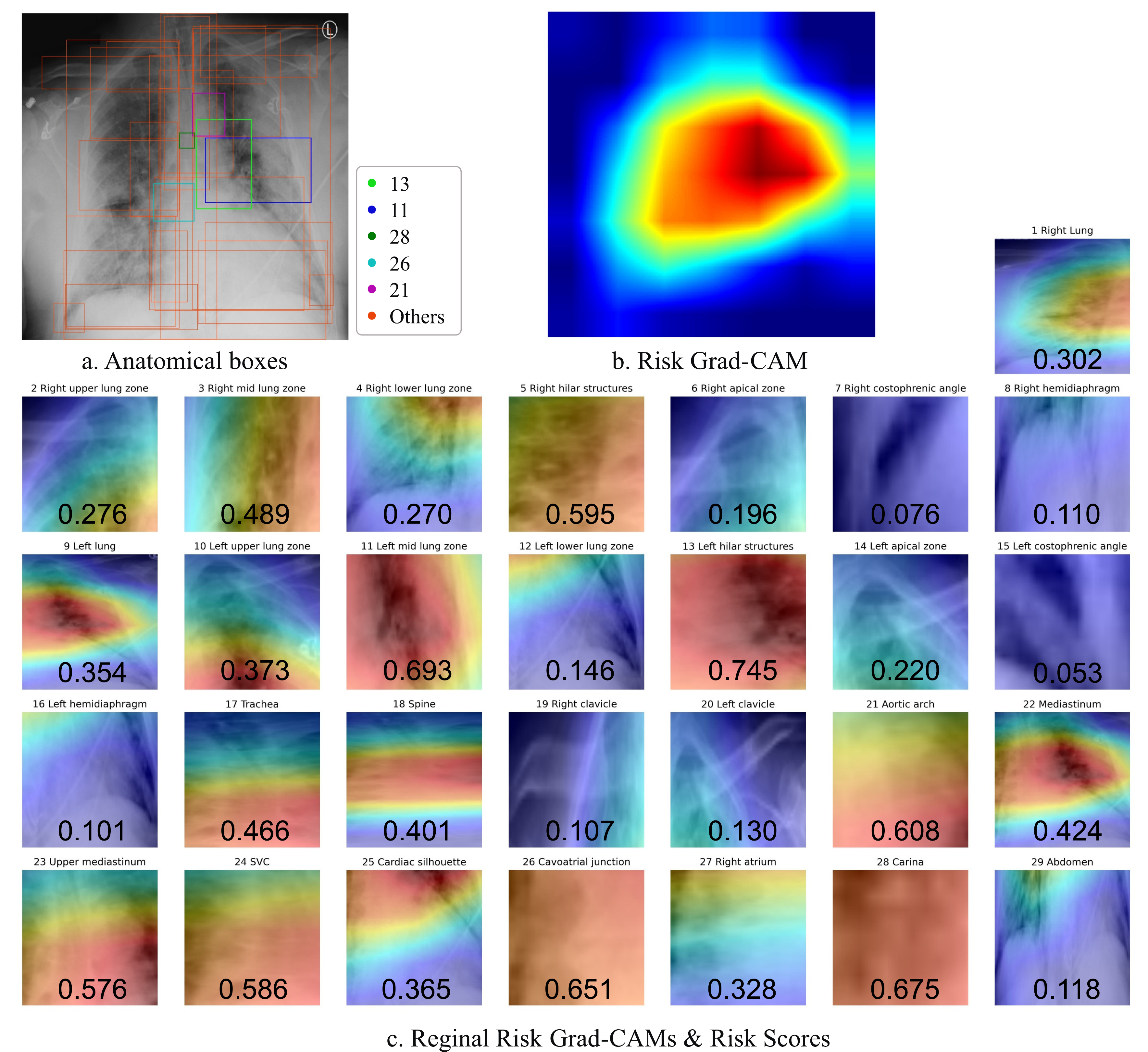}
\caption{Visualizations of the regional interpretable results. } \label{region}
\end{figure}

In terms of qualitative assessment, we have validated the regional interpretability of our model. The findings depicted in Fig. \ref{region} show a representative case from dataset B, providing insights into regional predictions, particularly emphasizing an CXR image featuring 29 predicted and annotated bounding boxes in subfigure a. Subfigure b presents risk-specific Grad-CAM based on the risk score from our survival prediction model, highlighting regions are the left lung, heart, and lower right lung. For this specific case, the model assigns a global risk score of 0.745. Through the analysis of anatomical region outcomes, we can delve into high-risk areas more comprehensively. \par
 
To enhance the visualization of regional highlights corresponding to local risks, subfigure c exhibits the risk Grad-CAM with each region delineated from global activation. The regional risk Grad-CAMs illustrate the prominence of the specific anatomical regions and reflecting local risk levels. Notably, regions such as 13, 11 and 26 exhibit deeper and broader red hues, indicating areas of disease attention. Correspondingly, top-5 bounding boxes in subfigure a reveal ground-glass opacities in the lung areas, a crucial diagnostic indicator for COVID-19. The interpretability of our model is evidenced through two avenues: (1) translating a single global risk score into regional highlights and risk values, thereby facilitating the comprehension of high-risk regions; (2) effectively portraying disease conditions through the ranked high-risk regions based on local risks, assisting clinicians in making informed decisions regarding diagnosis.

\section{Conclusions}
In this study, we presented a deep survival prediction model for COVID-19 risk stratification using CXR with an emphasis on interpretability to improve trustworthiness in clinical decision making. Integrating with large-scale pretrained knowledge, the efficiency and generalization ability were evaluated on a multi-center survival dataset, demonstrating via quantitative and qualitative assessments. Furthermore, our method produces regional outcomes provide transparency in elucidating the reasoning of disease attention to healthcare professionals. Our explainable deep survival prediction contributes to more reliable and transparent AI models by addressing concerns related to opacity and "black box" models that hinder acceptance and adoption in clinical practice. This research underscores importance of responsible AI development for trustworthy systems that ultimately benefit both clinicians and patients.


%
%
%
\bibliographystyle{splncs04}
\bibliography{ref}

\begin{thebibliography}{10}
\providecommand{\url}[1]{\texttt{#1}}
\providecommand{\urlprefix}{URL }
\providecommand{\doi}[1]{https://doi.org/#1}

\bibitem{cheng2023prior}
Cheng, P., Lin, L., Lyu, J., Huang, Y., Luo, W., Tang, X.: Prior: Prototype representation joint learning from medical images and reports. In: Proceedings of the IEEE/CVF International Conference on Computer Vision. pp. 21361--21371 (2023)

\bibitem{coxmodel}
Fox, J., Weisberg, S.: Cox proportional-hazards regression for survival data. An R and S-PLUS companion to applied regression  \textbf{2002} (2002)

\bibitem{gao2021risk}
Gao, Y.d., Ding, M., Dong, X., Zhang, J.j., Kursat~Azkur, A., Azkur, D., Gan, H., Sun, Y.l., Fu, W., Li, W., et~al.: Risk factors for severe and critically ill covid-19 patients: a review. Allergy  \textbf{76}(2),  428--455 (2021)

\bibitem{geis2019ethics}
Geis, J.R., Brady, A.P., Wu, C.C., Spencer, J., Ranschaert, E., Jaremko, J.L., Langer, S.G., Borondy~Kitts, A., Birch, J., Shields, W.F., et~al.: Ethics of artificial intelligence in radiology: summary of the joint european and north american multisociety statement. Radiology  \textbf{293}(2),  436--440 (2019)

\bibitem{goergen2013evidence}
Goergen, S.K., Pool, F.J., Turner, T.J., Grimm, J.E., Appleyard, M.N., Crock, C., Fahey, M.C., Fay, M.F., Ferris, N.J., Liew, S.M., et~al.: Evidence-based guideline for the written radiology report: Methods, recommendations and implementation challenges. Journal of medical imaging and radiation oncology  \textbf{57}(1), ~1--7 (2013)

\bibitem{guidotti2018survey}
Guidotti, R., Monreale, A., Ruggieri, S., Turini, F., Giannotti, F., Pedreschi, D.: A survey of methods for explaining black box models. ACM computing surveys (CSUR)  \textbf{51}(5),  1--42 (2018)

\bibitem{cindex}
Harrell~Jr, F.E., Lee, K.L., Califf, R.M., Pryor, D.B., Rosati, R.A.: Regression modelling strategies for improved prognostic prediction. Statistics in medicine  \textbf{3}(2),  143--152 (1984)

\bibitem{he2022masked}
He, K., Chen, X., Xie, S., Li, Y., Doll{\'a}r, P., Girshick, R.: Masked autoencoders are scalable vision learners. In: Proceedings of the IEEE/CVF conference on computer vision and pattern recognition. pp. 16000--16009 (2022)

\bibitem{hung2010estimation}
Hung, H., Chiang, C.T.: Estimation methods for time-dependent auc models with survival data. Canadian Journal of Statistics  \textbf{38}(1),  8--26 (2010)

\bibitem{rsf}
Ishwaran, H., Kogalur, U.B., Blackstone, E.H., Lauer, M.S.: Random survival forests  (2008)

\bibitem{johnson2019mimic}
Johnson, A.E., Pollard, T.J., Berkowitz, S.J., Greenbaum, N.R., Lungren, M.P., Deng, C.y., Mark, R.G., Horng, S.: Mimic-cxr, a de-identified publicly available database of chest radiographs with free-text reports. Scientific data  \textbf{6}(1), ~317 (2019)

\bibitem{km}
Kaplan, E.L., Meier, P.: Nonparametric estimation from incomplete observations. Journal of the American statistical association  \textbf{53}(282),  457--481 (1958)

\bibitem{deepsurv}
Katzman, J.L., Shaham, U., Cloninger, A., Bates, J., Jiang, T., Kluger, Y.: Deepsurv: personalized treatment recommender system using a cox proportional hazards deep neural network. BMC medical research methodology  \textbf{18}(1),  1--12 (2018)

\bibitem{kvamme2019time}
Kvamme, H., Borgan, {\O}., Scheel, I.: Time-to-event prediction with neural networks and cox regression. arXiv preprint arXiv:1907.00825  (2019)

\bibitem{ren2015faster}
Ren, S., He, K., Girshick, R., Sun, J.: Faster r-cnn: Towards real-time object detection with region proposal networks. Advances in neural information processing systems  \textbf{28} (2015)

\bibitem{rimmer2017radiologist}
Rimmer, A.: Radiologist shortage leaves patient care at risk, warns royal college. BMJ: British Medical Journal (Online)  \textbf{359} (2017)

\bibitem{selvaraju2017grad}
Selvaraju, R.R., Cogswell, M., Das, A., Vedantam, R., Parikh, D., Batra, D.: Grad-cam: Visual explanations from deep networks via gradient-based localization. In: Proceedings of the IEEE international conference on computer vision. pp. 618--626 (2017)

\bibitem{tanida2023interactive_rgrg}
Tanida, T., M{\"u}ller, P., Kaissis, G., Rueckert, D.: Interactive and explainable region-guided radiology report generation. In: Proceedings of the IEEE/CVF Conference on Computer Vision and Pattern Recognition. pp. 7433--7442 (2023)

\bibitem{wu2021chest}
Wu, J., Agu, N., Lourentzou, I., Sharma, A., Paguio, J.A., Yao, J.S., Dee, E.C., Kashyap, S., Giovannini, A., Celi, L.A., et~al.: Chest imagenome dataset for clinical reasoning. In: Annual Conference on Neural Information Processing Systems (2021)

\bibitem{zhou2016learning}
Zhou, B., Khosla, A., Lapedriza, A., Oliva, A., Torralba, A.: Learning deep features for discriminative localization. In: Proceedings of the IEEE conference on computer vision and pattern recognition. pp. 2921--2929 (2016)

\end{thebibliography}

\end{document}